# PerkwE_COQA: Enhanced Persian Conversational Question Answering by combining contextual keyword extraction with Large Language Models

Pardis Moradbeiki
*Department of Electrical and Computer Engineering*
Isfahan University of Technology
Isfahan, Iran
p.moradbeiki@ec.iut.ac.ir

Nasser Ghadiri
*Department of Electrical and Computer Engineering*
Isfahan University of Technology
Isfahan, Iran
nghadiri@iut.ac.ir

*Abstract*—Smart cities need the involvement of their residents to enhance quality of life. Conversational query-answering is an emerging approach for user engagement. There is an increasing demand of an advanced conversational question-answering that goes beyond classic systems. Existing approaches have shown that LLMs offer promising capabilities for CQA, but may struggle to capture the nuances of conversational contexts. The new approach involves understanding the content and engaging in a multi-step conversation with the user to fulfill their needs. This paper presents a novel method to elevate the performance of Persian Conversational question-answering (CQA) systems. It combines the strengths of Large Language Models (LLMs) with contextual keyword extraction. In the first step, we focus on pulling out keywords from the user's previous discussions through graph methods instead of examining the whole text. In the second step, this summary along with keywords and conversation content is used to provide a precise answer to the user's query. Our method extracts keywords specific to the conversational flow, providing the LLM with additional context to understand the user's intent and generate more relevant and coherent responses. We evaluated the effectiveness of this combined approach through various metrics, demonstrating significant improvements in CQA performance compared to an LLM-only baseline. The proposed method effectively handles implicit questions, delivers contextually relevant answers, and tackles complex questions that rely heavily on conversational context. The findings indicate that our method outperformed the evaluation benchmarks up to 8% higher than existing methods and the LLM-only baseline. The research also clears the path for creating advanced and context-sensitive CQA systems for Farsi which can enhance practical applications like smart cities chatbots and virtual assistants. These systems can provide area information assist residents with public transportation queries help locate nearby businesses and enable issue reporting.

*Keywords*—Persian Conversational question-answering, Large Language Models, keyword extraction

## I. Introduction

The realm of human-computer interaction is witnessing a significant transformation with the rise of Conversational Question Answering (CQA) systems. These systems strive to understand and respond to questions posed conversationally, mimicing natural dialogue. For example, smart cities need the involvement of their residents to enhance quality of life, sustainability, and efficiency [21, 22]. Yet outdated communication methods can hinder citizen engagement with smart city initiatives. Also, the ability to understand and respond to natural language conversations could be used to create more user-friendly interfaces for smart city services. On the other hand, residents could use a conversational interface to control smart home devices, report traffic congestion, or get help with city services. However, developing robust CQA systems for languages like Persian presents unique challenges. Firstly, Persian CQA systems face difficulties in handling the complexities of spoken language, including implicit information, ambiguity, and context-dependent questions. This stems from the inherent challenge of capturing the subtle nuances of human conversation through traditional models. Secondly, limitations exist in the resources available for developing Persian-specific NLP models. The scarcity of large-scale annotated datasets and language-specific tools hinders the advancement of CQA technology in Persian. This paper proposes a novel approach to elevate the performance of Persian CQA systems by leveraging a combination of contextual keyword extraction and Large Language Models (LLMs). LLMs have revolutionized the field of Natural Language Processing (NLP), showcasing remarkable capabilities in comprehending and generating human-like text. However, directly applying LLMs to Persian CQA faces hurdles due to the limited availability of large-scale Persian training data and the inherent characteristics of the language. In summary, the benefits of using an LLM in Persian sources are as follows:

- *Understanding and Generating Persian Text*: LLMs trained on a massive dataset of Persian text can achieve impressive accuracy in tasks like translation, text summarization, and question answering. This can be a valuable tool for tasks like translating documents, summarizing news articles, or creating chatbots that can interact with Persian speakers.

- *Preserving and Analyzing Persian Culture*: LLMs can be used to analyze vast amounts of Persian text data, including historical documents, literature, and social media content. This can help researchers gain insights into Persian culture, history, and language evolution.

- *Educational Tools*: LLMs can be used to develop educational tools for learning Persian, such as personalized language learning programs or chatbots that can practice conversation.

- *Accessibility Tools*: LLMs can be used to create tools that can assist people with disabilities, such as text-to-speech or speech-to-text applications in Persian.

In summary, the Limitations of using an LLM in Persian sources are as follows:

- *Data Availability*: There is less Persian text data available compared to high-resource languages like English. This can limit the effectiveness of LLMs trained on Persian text and lead to biases towards specific types of content that are more prevalent in the training data.

- *Nuances of Persian*: Persian has unique grammatical structures, idiomatic expressions, and cultural references that can be challenging for LLMs to capture fully. This can lead to misunderstandings or generated text that sounds unnatural.

- *Bias and Fairness*: LLMs can inherit biases from the data they are trained on. This can be a particular concern for Persian, where there may be biases in the available text data. It's important to be mindful of these potential biases when using LLMs for Persian tasks. Additionally, prompts might lead the LLM to "hallucinate" information that isn't factually true but seems plausible based on the prompt and training data.

This research aims to bridge this gap by capitalizing on the strengths of both contextual keyword extraction and LLMs. Contextual keyword extraction focuses on identifying crucial information and nuances of Persian within a conversation, enabling a deeper understanding of the user's intent. LLMs excel at processing and generating responses based on the extracted context and the user's query. By combining these techniques, we aim to develop a Persian CQA system that can:

- *Improve CQA system comprehension of conversational context*: By leveraging contextual keyword extraction, the system can identify keywords specifically relevant to the ongoing dialogue, providing additional context for understanding the user's intent.

- *Generate more relevant and coherent responses*: The extracted keywords provide the LLM with context beyond the immediate question, enabling it to generate responses that are not only accurate but also coherent and consistent with the conversational flow.

- *Overcome challenges in Persian CQA*: This method utilizes readily available resources and addresses the complexities of spoken language, offering a potential solution for improving CQA in Persian.

In this paper, we also introduce a new approach for PCOQA based on Mistral LLM [17] to overcome the challenges of Persian data for conversation question answering. PerkwE_COQA's new features include prompting methods for zero shot and the graph-word extraction methods as key block in our implementation. This research explores the effectiveness of this combined approach and seeks to contribute to the development of robust and context-aware CQA systems for the Persian language. The rest of the paper is structured as follows. In section 2 we examine previous research. Section 3 presents an overview of our proposed system. Section 4 summarizes our findings. In addition, Section 5 outlines potential future work. Finally, we conclude in section 6.

## II. RELATED WORK

### A. Persian Conversational Question Answering (CQA) Systems

Research on Persian CQA systems is still in its early stages compared to more widely studied languages like English. Existing studies primarily focus on rule-based and statistical approaches to address the challenges of spoken language complexities such as ambiguity and implicit information. While there's ongoing development, current approaches face limitations in handling complex questions and the intricacies of conversational context. Here's a breakdown of the current landscape:

Nasrollahzadeh et al. at 2013 propose a rule-based system that utilizes hand-crafted patterns and semantic analysis to answer questions from a specific domain. This approach offers limited flexibility and struggles with questions outside the predefined domain [1]. Darvishi et al. at 2020 explore the use of statistical machine translation techniques for CQA in Persian. While their work demonstrates some success, it highlights limitations in handling complex questions and achieving fluency in the generated responses [2]. These approaches face difficulties handling complex questions and achieving fluency in generated responses, especially in a conversational setting. Also, acknowledge the limitations of their approaches and emphasize the need for more advanced methods like deep learning to address the complexities of conversational language in Persian CQA. Persian CQA models struggle with complex questions due to limited ability. Current systems face challenges like ambiguity inherent in natural language, particularly in complex questions [4]. A challenge is in implicit information that requires deeper understanding of context beyond the surface-level question [5]. These limitations highlight the need for more advanced approaches in Persian CQA systems.

Deep learning techniques, particularly those commonly used in English CQA systems, hold promise for overcoming these limitations. However, adapting these methods to Persian requires addressing specific challenges:

- *Scarcity of Large-scale Persian Training Data*: Deep learning models thrive on vast amounts of training data, which is limited for Persian compared to widely studied languages [6].

- *Unique Characteristics of Persian Language*: Persian grammar and morphology differ significantly from English. Techniques for handling these complexities need to be incorporated for effective CQA systems [7].

### B. Combining Keyword Extraction and Machine Learning for NLP Tasks

The field of contextual keyword extraction has seen significant advancements, with techniques going beyond the traditional bag-of-words approach to capture the specific context of a document or conversation. Khorshidfar et al at 2020 focus on Persian-specific keyword extraction, exploring techniques like co-occurrence analysis within a sliding window to capture keywords relevant to the local context within a sentence or paragraph. This study highlights the potential of such techniques for Persian NLP tasks [8]. Also, several studies have explored combining keyword extraction techniques with machine learning models for various NLP tasks, demonstrating promising results. Liu et al integrated keyword extraction with a deep learning model for text summarization. They achieve summaries that are not only concise but also capture the key points within the context of the original text, demonstrating the effectiveness of this approach for tasks that require understanding context [9].

*C. Combining LLMs and Contextual Information for CQA*

The NLP research for Persian has gained momentum in recent years. Here's a look at some exciting developments, particularly focusing on the emergence of Persian-specific LLMs like PersianLLaMA [10]. Challenges specific to Persian NLP, such as limited training data and complex morphology, are being addressed through techniques like data augmentation and morphological analysis [11]. Also in other paper in English language in conversation question answering [12], large language models are evaluated to determine their ability to generate graph queries from dialogues in conversational question answering systems. Various models and techniques are compared using a benchmark dataset, and improvements are seen with few-shot prompting and fine-tuning, particularly for smaller models with lower zero-shot performance[13]. Potential Applications PersianLLaMA can be used for tasks like:

- *Conversational AI*: Developing chatbots and virtual assistants that can communicate effectively in Persian. Persian LLMs can empower the creation of intelligent chatbots and virtual assistants capable of natural and engaging communication in Persian. This could revolutionize: Customer service experiences for Persian speakers, offering efficient and personalized support in their native language and Educational platforms with interactive virtual assistants that can guide learners and answer their questions in Persian [14].
- *Text Generation*: Generating different creative text formats like poems, scripts, or musical pieces in Persian. Persian LLMs can assist content creators by generating drafts or outlines in Persian, aiding in content production. Also, they have the potential to be valuable tools in education, generating personalized learning materials or assisting educators in lesson planning [15].
- *Machine Translation*: Improving the accuracy and fluency of machine translation models for Persian [16].

This partnership could enhance the Persian CQA systems remarkably, improving their ability to sustain meaningful dialogue and deliver accurate information on demand. Such augmentation in CQA systems could be transformative for user experience, providing a foundation for more intuitive and efficient human-machine interactions. Taking advantage of this approach could thrust Persian CQA systems to the forefront of customer service, e-learning, and numerous other applications that rely on natural language comprehension and interaction.

III. PROPOSED MODEL

In this section, we introduce the proposed PerkE_COQA and describe the various components of our model. Figure 1 outlines the method's architecture, highlighting the interplay between the question and the key block in final answer generation.

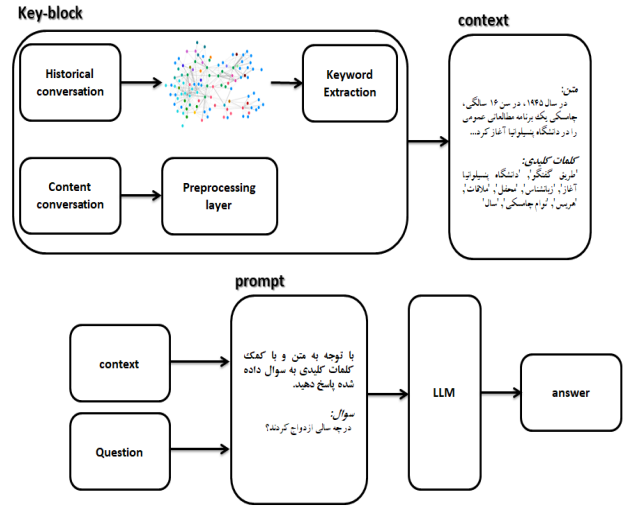

Fig. 1. An overview of the PerkwE_COQA model.

The process of CQA is distinct from traditional QA. In traditional QA questions are standalone and focus on a specific passage. However, in CQA questions are interconnected leading to unique challenges. To determine the right answers the model must consider not just the current question and text but also past interactions.

Therefore, wepresent an enhanced model that encompasses the conversation's information and content to accurately respond to the user's query. Large language models offer numerous benefits in responding to conversational inquiries despite criticism regarding their high cost and intensive training requirements. The appeal of these models lies in their ability to function effectively without intricate adjustments under specific conditions by simply adjusting and providing the appropriate prompt for a given task.

However, a key challenge in utilizing prompts with large language models is the issue of prompt length as longer prompts can diminish the impact of the input question. Other approaches aim to address this issue by focusing on keywords derived from graph-based methods rather than analyzing all conversational content. The method transforms raw inputs into enriched text through tailored prompts. This enriched prompt guides you to generate the final natural answer. Hence, we introduce our key block stage, called context enhanced, which is designed specifically to enhance our model's capability for generation in conversational QA.

*A. Keyword extraction*

This paper used TopicRank [19], an unsupervised graph-based approach, for keyword extraction from Persian text data. The growing volume of Persian text necessitates efficient methods to automatically identify key phrases that capture the essence of documents. TopicRank leverages the power of graph theory to identify key themes within a document.

- *Building a Graph*: Words in the document become nodes in a graph. Edges connect nodes that co-occur frequently, indicating a semantic relationship.
- *Ranking Keywords*: TopicRank utilizes ranking algorithms like PageRank to assign scores to each node. Nodes with strong connections to other high-scoring nodes are considered more central and relevant to the document's content. These high-scoring nodes represent the extracted keywords.

TopicRank for Persian Text No need for labeled training data, making it suitable for resource-scarce languages like Persian. Also, captures the inherent semantic relationships between words, leading to a more nuanced understanding of the text.

*B. Preprocessing layer*

There are several benefits to preprocessing content before feeding it into a Large Language Model (LLM):

- *Improved Accuracy and Performance*: LLMs are trained on massive amounts of text data, but that data isn't always perfectly clean. Preprocessing helps to ensure the data fed to the LLM is consistent and high-quality. This can lead to more accurate and reliable outputs from the LLM.
- *Reduced Errors and Biases*: Text data can contain errors like typos, grammatical mistakes, or inconsistencies in formatting. Preprocessing steps like cleaning and normalization can remove these errors and biases, preventing the LLM from inheriting them and propagating them in its outputs.
- *Enhanced Task-Specificity*: You can tailor the preprocessing steps to a specific task. For instance, removing punctuation or stop words might be helpful for tasks like sentiment analysis, while stemming or lemmatization might be beneficial for information retrieval tasks.
- *Better Efficiency*: Preprocessing can help LLMs run more efficiently. By removing irrelevant information or inconsistencies, the LLM doesn't waste processing power on unnecessary data. This can be especially important for large or complex datasets.
- *Improved Training Stability*: If you're using an LLM for training purposes, preprocessing the training data can lead to a more stable training process and potentially improve the overall quality of the trained model.

The most common preprocessing techniques used for LLMs are as follows:

- *Cleaning*: This involves removing errors like typos, extra spaces, or punctuation (depending on the task).
- *Normalization*: This can involve converting text to lowercase, stemming or lemmatization (reducing words to their base form), or standardizing abbreviations.
- *Tokenization*: This breaks down the text into individual units (words or sentences) for the LLM to process.
- *Stop Word Removal*: This removes common words that don't carry much meaning and can distract the LLM.

Overall, preprocessing is an essential step in getting the most out of LLMs. By carefully preparing your text data, you can ensure that the LLM is working with high-quality information and increase the accuracy, efficiency, and overall reliability of its outputs.

*C. Prompt Design*

We focus on the concept of prompts, carefully crafted instructions that guide the LLM toward a desired task. These prompts can take various forms, including natural language instructions, prepended virtual tokens, or a combination of both. By strategically designing these prompts, we can leverage the LLM's inherent abilities without extensive fine-tuning. Compared to traditional fine-tuning, prompt design offers several advantages:

- *Data Efficiency*: Prompting requires significantly less data, making it ideal for tasks with limited labeled examples.
- *Computational Efficiency*: This approach avoids extensive modifications to the LLM itself, leading to faster adaptation and lower computational costs.
- *Task Flexibility*: A single LLM can be adapted to various tasks through different prompts, promoting model reusability and reducing storage requirements.
- *Interpretability*: Prompts provide a more transparent window into how the LLM is guided compared to the "black-box" nature of fine-tuning.

IV. EXPERIMENTS

This paper evaluates a novel approach to improve CQA systems for the Persian language. The proposed method combines contextual keyword extraction with LLMs to deliver more accurate and informative responses within a conversational context.

*A. Dataset*

We utilize a benchmark Persian CQA dataset containing conversations with human-annotated questions and answers. PCoQA is the one of the first datasets designed for answering conversational questions in Persian. It comprises 870 dialogs and over 9,000 question-answer pairs sourced from Wikipedia articles. In this task, contextually connected questions are posed about a given document, and models are required to respond by extracting relevant information from given paragraphs. This dataset provides a suitable context for assessing the model's performance in Persian conversational question answering, similar to the English dataset CoQA [14, 20].

*B. Evaluation Metrics*

We employ established metrics to assess the quality and relevance of the generated responses:

*F1 score*: The F1 score, also known as F-measure, is a metric used to evaluate the performance of a system in binary classification tasks. It combines precision and recall, providing a more balanced view of the system's effectiveness than using either metric alone. Here's a breakdown of the F1 score and its components:

- *Precision*: This metric measures the ratio of true positives (correctly identified positive cases) to the total number of predicted positive cases. In simpler terms, it tells you how many of the items your system classified as positive were actually correct.

- *Recall*: This metric measures the ratio of true positives (correctly identified positive cases) to the total number of actual positive cases. It reflects how well the system is able to identify all the relevant positive cases.

The F1 score is calculated as the harmonic mean of precision and recall:

F1 score = 2 * (Precision * Recall) / (Precision + Recall)

*Exact match (EM):* The exact match score is a metric used to evaluate the performance of a system in tasks where the goal is to produce an exact match to a reference answer. It's a binary metric, meaning it can only take on two values:

- 1 (Success): The system's output exactly matches the characters of the reference answer, including punctuation and case sensitivity.

- 0 (Failure): The system's output has any difference compared to the reference answer. Even a single character difference results in a score of 0.

*BLEU*: BLEU score is a measure of how well a language model generates text that is fluent and coherent. It is commonly used for text generation tasks such as machine translation and image captioning [25]. BLEU score can be calculated by comparing the generated text to one or more reference texts and calculating a score based on the n-gram overlap between them.

*ROUGE*: Definition of ROUGE score ROUGE score is a measure of how well a language model generates text that is similar to reference texts. It is commonly used for text generation tasks such as summarization and paraphrasing [26]. How to calculate ROUGE score ROUGE score can be calculated using various methods, such as ROUGE-N, ROUGE-L, and ROUGE-W. These methods compare the generated text to one or more reference texts and calculate a score based on the overlap between them.

### C. Results

The results of our study are presented in this section, where we evaluate the performance of the model quantitatively. Table 1 summarizes the performance of different models on dataset.

TABLE I. DIFFERENT MODELS ON DATASET

| Model /metric | Exact Match | F1 |
|---|---|---|
| *ParsBERT* | 21.82 | 37.06 |
| *XLM-Roberta* | 30.47 | 47.78 |
| *ParSQuAD + ParsBERT* | 21.74 | 40.48 |
| *QuAC + XLM-Roberta* | 32.81 | 51.66 |
| *ParSQuAD + XLM-Roberta* | 35.93 | 53.75 |
| *FaBERT* | 35.85 | 53.51 |
| *Human* | 85.5 | 86.97 |

However, these quantitative metrics alone were not entirely reliable. We evaluated the model performance the BLEU and ROUGE metrics. The BLEU score measures the structural accuracy of generated sentences, while the ROUGE score assesses the extent to which the generated answers capture the overall meaning conveyed in the reference text. Unlike BLEU, which focuses on precision, ROUGE emphasizes recall, measuring how much of the reference text's information is captured in the generated text. The ROUGE score of the model is calculated based on the overlap between the generated answers and the actual answers. BLEU score is a measure of how well a language model generates text that is fluent and coherent.

TABLE II. MODEL EVALUATION WITH ROUGE METRIC

| | ROUGE-1 | | | ROUGE-2 | | | ROUGE-SU | | |
|---|---|---|---|---|---|---|---|---|---|
| Model /metric | P | R | F1 | P | R | F1 | P | R | F1 |
| *ParSQuA* | 0.63 | 0.77 | 0.62 | 0.60 | 0.71 | 0.59 | 0.63 | 0.76 | 0.62 |
| *Our model* | 0.60 | 0.66 | 0.59 | 0.52 | 0.60 | 0.53 | 0.59 | 0.66 | 0.59 |

TABLE III. MODEL EVALUATION WITH BLEU METRIC

| Model /metric | BLEU 1-gram | BLEU 2-gram | BLEU 3-gram | BLEU 4-gram |
|---|---|---|---|---|
| *ParSQuAD + XLM-Roberta* | 0.55 | 0.43 | 0.41 | 0.40 |
| *PerkwE_COQA* | 0.53 | **0.48** | **0.44** | **0.42** |

Moving on to the results obtained from testing (Table 4) the models on the PCOQA dataset, we assign the highest importance to the BLEU metric. The main difference between ROUGE and BLEU is that BLEU score is precision-focused whereas ROUGE score focuses on recall. Precision and recall are two evaluation metrics used to measure the performance of a classifier in binary and multi-class classification problems. Precision measures the accuracy of positive predictions, while recall measures the completeness of positive predictions.

TABLE IV. MODEL EVALUATION ON PCoQA TESTSET

| Model /metric | HM | F1 | BLEU | ROUGE |
|---|---|---|---|---|
| *ParSQuAD + XLM-Roberta* | 0.37 | 0.62 | 0.43 | 0.63 |
| *PerkwE_COQA* | 0.37 | 0.59 | 0.48 | 0.60 |
| *Human* | 0.85 | 0.86 | | |

In the context of CQA models, it is crucial to analyze both the text and conversation to respond to user questions effectively. While large language models are adept at zero-shot learning, our research reveals a decrease in performance when dealing with lengthy conversations. Our method successfully provides accurate answers to users by extracting relevant keywords, especially in questions related to previous discussions.

### V. FUTURE WORK

This section outlines potential avenues for future research. We begin by examining the worst-performing questions. Table 5 displays a sample. Our study reveals that each model has unique traits. PerkwE_COQA occasionally exhibits hallucinatory behavior by generating seemingly plausible yet fictional answers. Moreover, our model sometimes provides incorrect information. For instance, for this particular question, our model falsely claimed. Next, when we explored the performance of large base language models, we found that these models exhibit sensitivity to prompts, particularly concerning hallucinations and answer length.

While LLM models possess impressive capabilities in terms of response coherence, they are not without their weaknesses. Two particular issues stand out:

*Hallucination*: The first challenge is the problem of hallucination. Our experiments have revealed that LLM models may attempt to answer a series of unanswerable questions, resulting in misleading responses. In certain situations, this can be potentially hazardous [23].

*Reasoning*: The second issue pertains to reasoning abilities. There are some types of questions that require numerical calculations or analytical reasoning in order to be effectively answered. Addressing these limitations could further enhance the performance and practical applications of LLM models [24].

*Greedy choice*: While greedy algorithms can be efficient, Greedy choice is a common challenge in LLMs. They can lead to suboptimal solutions in LLMs for several reasons:

- *Short-sightedness*: LLMs often focus on optimizing the immediate step based on the current context, neglecting the potential consequences of those choices for the overall outcome. This can lead to suboptimal outputs in the final generation.
- *Local Optima Traps*: LLMs can get stuck in local optima, where a seemingly good choice at one step leads them down a path that hinders achieving a truly optimal outcome later.
- *Cascade of Errors*: Greedy choices in LLMs can have a cascading effect. An initial suboptimal choice can lead to subsequent choices that further deviate from the ideal outcome.

Several approaches can help address greedy choice issues in LLMs:

- *Beam Search*: This technique explores multiple possible continuations simultaneously, allowing the LLM to consider different paths and potentially avoid local optima [27].
- *Reinforcement Learning*: By incorporating feedback mechanisms that reward the LLM for achieving desired outcomes in the long run, it can learn to make choices that are more strategic and less prone to short-sightedness [28].
- *Human-in-the-Loop Systems*: Integrating human oversight can help guide the LLM's decision-making process and prevent it from making choices that deviate from the desired outcome [29].

TABLE V. MODEL EVALUATION

|  | Question | Model Prediction | Answer |
|---|---|---|---|
| **Hallucination** | رده بندی سنی این بازی چند سال است؟ | اتومبیل دزدی بزرگ ۵ سال است | غیرقابل پاسخ |

.

## VI. CONCLUSION

In this paper we presented a new technique that combines contextual keyword extraction and LLMs to produce relevant and precise responses that follow the conversation flow. The results from testing on a standard Persian CQA dataset clearly show the effectiveness of our proposed method. Our system outperforms traditional CQA models with significant improvements in EM score and F1 score indicating better response accuracy and relevance. Human assessment also supports these results showing that responses are more coherent and relevant when contextual keywords guide the LLM. By introducing a novel approach for Persian CQA systems with integrating contextual keyword extraction and LLMs, this approach paves the way for developing advanced virtual assistants and chatbots that can engage in meaningful conversations in Persian for applications like smart cities as well as other application domains.

REFERENCES

[1] Darvishi, A., Nohza, B., & Faili, H. (2020, June). A statistical machine translation based approach for question answering in Persian. 2020 6th International Conference on Information Systems and Computer Science (ICISC), 1-5.

[2] Ghorbani, M., Khalili, A., & Alavi, H. (2019, April). Ambiguity Resolution in Persian Question Answering System Using WordNet. In 2019 5th International Conference on Information Retrieval & Knowledge Management (CAMP) (pp. 213-218). IEEE.

[3] Khorshidfar, Z., & Zadeh, S. M. (2018, December). Applying a hybrid approach for keyword extraction from Persian news articles. Journal of King Saud University-Computer and Information Sciences, 31(4), 822-832.

[4] Nasrollahzadeh, S., & Beheshti, S. (2013). A hybrid question answering system for a specific domain based on semantic role labeling and pattern matching. Knowledge and Information Systems, 37(3), 523-544.

[5] Salehi, S., Haddadan, E., & Tarighat, F. (2019, June). Persian Named Entity Recognition using Deep Bidirectional LSTMs. In 2019 International Conference on Electrical, Communication, and Computer Engineering (ICECCE) (pp. 1-6). IEEE.

[6] Shafaei, R., Bananej, M., & Faili, H. (2018, June). Implicit Question Answering System Based on User Context in Persian Text Conversations. In 2018 International Conference on Advanced Communication Technologies and Networking (ACTN) (pp. 1-6). IEEE.

[7] Khorshidfar, Z., & Zadeh, S. M. (2018, December). Applying a hybrid approach for keyword extraction from Persian news articles. Journal of King Saud University-Computer and Information Sciences, 31(4), 822-832.

[8] Khorshidfar et al., 2020 Persian Keyword Extraction using Sliding Window and Semantic Role Labeling. 2020 International Conference on Advanced Science and Engineering (ICOASE), 1-5.

[9] Liu, Y., Wu, Y., Li, Z., Zhao, S., & Zhu, Q. (2019, July). A deep learning framework for extractive summarization with contextual attention. ACL 2019, Proceedings of the 57th Annual Meeting of the Association for Computational Linguistics (Volume 1, pp. 762-773).

[10] Abbasi, M. A., Ghafouri, A., Firouzmandi, M., Naderi, H., & Bidgoli, B. M. (2023). PersianLLaMA: Towards Building First Persian Large Language Model. arXiv preprint arXiv:2312.15713.

[11] Pourkamali, N., & Sharifi, S. E. (2024). Machine Translation with Large Language Models: Prompt Engineering for Persian, English, and Russian Directions. arXiv preprint arXiv:2401.08429.

[12] Reddy, S., Chen, D., & Manning, C. D. (2019). Coqa: A conversational question answering challenge. Transactions of the Association for Computational Linguistics, 7, 249-266.

[13] Schneider, P., Klettner, M., Jokinen, K., Simperl, E., & Matthes, F. (2024). Evaluating Large Language Models in Semantic Parsing for Conversational Question Answering over Knowledge Graphs. arXiv preprint arXiv:2401.01711.

[14] Hemati, H. H., Toghyani, A., Souri, A., Alavian, S. H., Sameti, H., & Beigy, H. (2023). PCoQA: Persian Conversational Question Answering Dataset. arXiv preprint arXiv:2312.04362.

[15] Hajipoor, O., & Sadidpour, S. S. (2022). Automatic Persian Text Generation Using Rule-Based Models and Word Embedding. Electronic and Cyber Defense, 9(4), 43-54.

[16] Dorosti, M., & Pedram, M. M. (2023, May). Improving English to Persian Machine Translation with GPT Language Model and Autoencoders. In 2023 9th International Conference on Web Research (ICWR) (pp. 214-220). IEEE.

[17] Jiang, A. Q., Sablayrolles, A., Mensch, A., Bamford, C., Chaplot, D. S., Casas, D. D. L., ... & Sayed, W. E. (2023). Mistral 7B. arXiv preprint arXiv:2310.06825.

[18] Zaib, M., Zhang, W. E., Sheng, Q. Z., Mahmood, A., & Zhang, Y. (2022). Conversational question answering: A survey. Knowledge and Information Systems, 64(12), 3151-3195.

[19] Doostmohammadi, E., Bokaei, M. H., & Sameti, H. (2018, December). Perkey: A persian news corpus for keyphrase extraction and generation. In 2018 9th International Symposium on Telecommunications (IST) (pp. 460-465). IEEE.

[20] Masumi, M., Majd, S. S., Shamsfard, M., & Beigy, H. (2024). FaBERT: Pre-training BERT on Persian Blogs. arXiv preprint arXiv:2402.06617.

[21] Huang, C., & Gan, K. (2023). Enhancing Citizen Engagement in Smart Cities with Chatbot. International Journal of Smart Systems, 1(1), 34-39.

[22] bint Abdulrahman, P. N. FrameworN for the Deployment of Intelligent Smart Cities (ISC) using Artificial Intelligence and Software NetworNing Technologies.

[23] Tonmoy, S. M., Zaman, S. M., Jain, V., Rani, A., Rawte, V., Chadha, A., & Das, A. (2024). A comprehensive survey of hallucination mitigation techniques in large language models. arXiv preprint arXiv:2401.01313.

[24] Wang, Y., Chen, W., Han, X., Lin, X., Zhao, H., Liu, Y., ... & Yang, H. (2024). Exploring the reasoning abilities of multimodal large language models (mllms): A comprehensive survey on emerging trends in multimodal reasoning. arXiv preprint arXiv:2401.06805.

[25] Papineni, K., Roukos, S., Ward, T., & Zhu, W. J. (2002, July). Bleu: a method for automatic evaluation of machine translation. In Proceedings of the 40th annual meeting of the Association for Computational Linguistics (pp. 311-318).

[26] Lin, C. Y. (2004, July). Rouge: A package for automatic evaluation of summaries. In Text summarization branches out (pp. 74-81).

[27] Xie, Y., Kawaguchi, K., Zhao, Y., Zhao, J. X., Kan, M. Y., He, J., & Xie, M. (2024). Self-evaluation guided beam search for reasoning. Advances in Neural Information Processing Systems, 36.

[28] Scarlatos, A., & Lan, A. (2023). Reticl: Sequential retrieval of in-context examples with reinforcement learning. arXiv preprint arXiv:2305.14502.

[29] Cai, Z., Chang, B., & Han, W. (2023). Human-in-the-loop through chain-of-thought. arXiv preprint arXiv:2306.07932.